\documentclass{article}
\usepackage{spconf,graphicx}
\usepackage{xcolor}
\usepackage{amsmath}
\usepackage{float}
\usepackage{comment}
\usepackage{subfig}
\usepackage{booktabs}  
\usepackage{makecell}
\usepackage{hyperref}
\usepackage{spconf}
\usepackage{float}
\usepackage{subfig}
\usepackage{tabularx}
\usepackage{lipsum}

\title{Region Extraction Based Approach For Cigarette Usage Classification Using Deep Learning}
\name{Anshul Pundhir*, Deepak Verma*, Puneet Kumar, Balasubramanian Raman\thanks{*Denotes Equal Contribution}}
\address{Computer Science and Engg. Dept., Indian Institute of Technology, Roorkee, India, 247667}

\begin{document}
	%
	\maketitle
	\begin{abstract}
		This paper has proposed a novel approach to classify the subjects' smoking behavior by extracting relevant regions from a given image using deep learning. After the classification, we have proposed a conditional detection module based on Yolo-v3, which improves model's performance and reduces its complexity. As per the best of our knowledge, we are the first to work on this dataset. This dataset contains a total of 2,400 images that include smokers and non-smokers equally in various environmental settings. We have evaluated the proposed approach's performance using quantitative and qualitative measures, which confirms its effectiveness in challenging situations. The proposed approach has achieved a classification accuracy of 96.74\% on this dataset.
	\end{abstract}
	\section{Introduction}\label{sec:intro}
	\begin{figure*}[]
		\centering
		\includegraphics[width=0.97\textwidth, height=7.3cm]{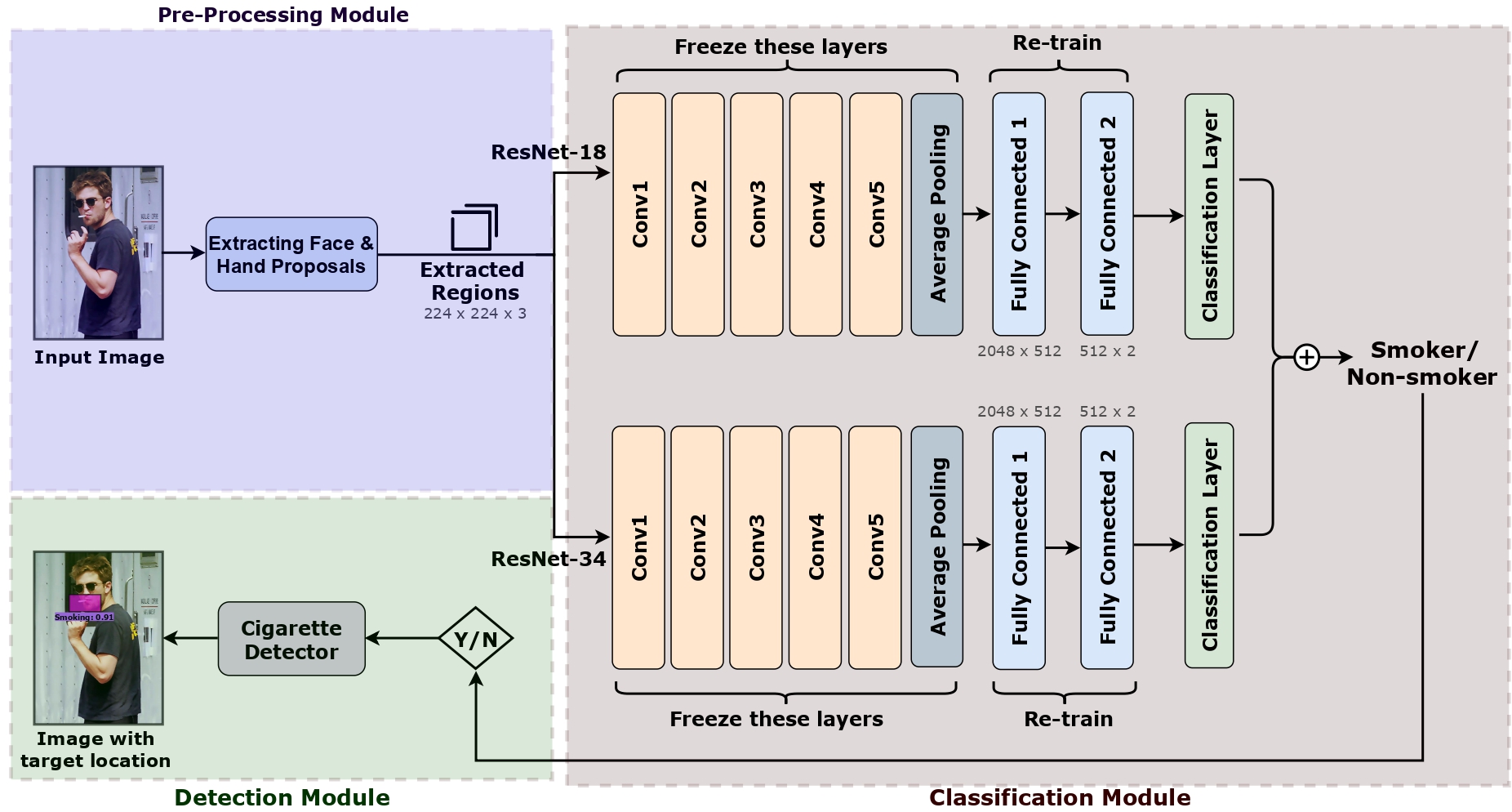}
		\caption{Schematic architecture of the proposed approach.\vspace{-.1in}}
		\label{fig:visina8}
	\end{figure*}
	
	Today's world, which is developing posthaste, has seen various technological innovations and financial advancements that positively serve society in many areas. Despite that, we have many problems such as pollution, an increasing number of road-accidents, health issues such as lung cancer, respiratory diseases, and eye-vision problems. These hazards happen due to various factors, out of which daily use of cigarettes is a prominent one. As per the doctors' advice, one should avoid cigarette use since it has adverse effects on our health, environment, and life span. The governments have also established the rules to avoid their use in public areas, but some break laws when they find themselves not monitored by any authority. Unfortunately, cigarette use by one person has adverse effects on others' lives in the form of pollution, health issues, and car accidents. So, there is an essential need to develop an automated system that can help to find a person's smoking behavior. Such systems have a wide range of applications such as automated smoke monitoring systems, cigarette censoring in videos, and controlling the number of road accidents due to drivers' smoking behavior~\cite{e08}.
	
	Various research attempts have been made for cigarette usage detection and classification~\cite{r01,r13,e01,e02,e03,e07,x1,x2,x4}. The existing works in this context are either image-based or sensor-based. The image-based methods process the image-related information such as the presence of smoke and the color of smoking object~\cite{e08,r01,r13,e02,e07,r02}. On the other hand, the sensor-based methods deploy the sensors to detect the smoking behavior and process the data collected by them~\cite{e01,e03,e04,e05,e06}. The literature survey suggests that this problem needs further exploration since most researchers have focussed on individual-level surroundings. Moreover, the problem includes challenges due to the tiny shape of a cigarette. Better cigarette usage detection and classification methods capable of addressing these challenges will contribute towards a safer and greener world.  It motivated us to develop an approach that can effectively overcome the aforementioned challenges and accurately detect smoking behavior.
	
	The proposed approach consists of a region extraction module, classification module, and conditionally active Yolo-v3~\cite{r10} based real-time detection module. It provides a simple yet effective tool to judge the subjects' smoking behavior by analyzing their visual information. The region extraction module refines the visual information by extracting face and hand proposals. The classification module processes these proposals for final classification. Based on the classification result, the detection module will perform cigarette detection. 
	
	The proposed approach has been evaluated on a recent dataset named `Dataset containing smoking and not-smoking images (smoker vs. non-smoker)'~\cite{dataset} containing 2,400 images with a nearly equal number of smoker and non-smoker images and achieved accuracy of 96.74\%. This approach's effectiveness is determined quantitatively by measuring accuracy, precision, recall, and qualitatively by visualizing its performance in different challenging situations. During the evaluation, results show that the proposed approach can handle various challenges like variability in hand, face postures, different illumination conditions, and little difference between smoker and non-smoker in the larger scene. The code for the work presented in this paper is available at \href{https://github.com/MIntelligence-Group/CigDetect}
	{\underline {https://github.com/MIntelligence-Group/CigDetect}}.\vspace{.05in}
	
	The contributions of the paper are summarised as follows. 
	\begin{itemize}\vspace{-.12in}
		\item A novel region extraction based deep-learning approach has been proposed for cigarette usage classification. It is capable of handling challenging situations such as low brightness, little visibility of cigarettes, and various gestures of hands.\vspace{-.06in}
		\item The incorporation of conditionally active detection has been observed to save the computational cost and improve the detection performance by reducing the false positives.\vspace{-.06in}
		\item The classification accuracy results obtained for various baseline models formulated during the ablation study verify the proposed approach's effectiveness. The proposed approach has obtained better results than the state-of-the-art methods for similar problems of identifying small objects in images.	
	\end{itemize} 
	
	The rest of the paper is organized as follows. The proposed methodology is discussed in section 2; experiments and results in section 3. Finally, in section 4, we have concluded our findings with future directions for further research. 
	
	\section{PROPOSED METHODOLOGY} \label{sec:format}
	This section elaborates on the proposed methodology. The proposed method's architecture has been shown in Fig.~\ref{fig:visina8}, and various components are discussed in the following sections. \vspace{-.04in}	
	
	\subsection{Region Extraction Module}
\begin{figure}[]%
		\centering
		\subfloat[\centering Faced bounding box]{{\includegraphics[width=2cm,height=2.5cm]{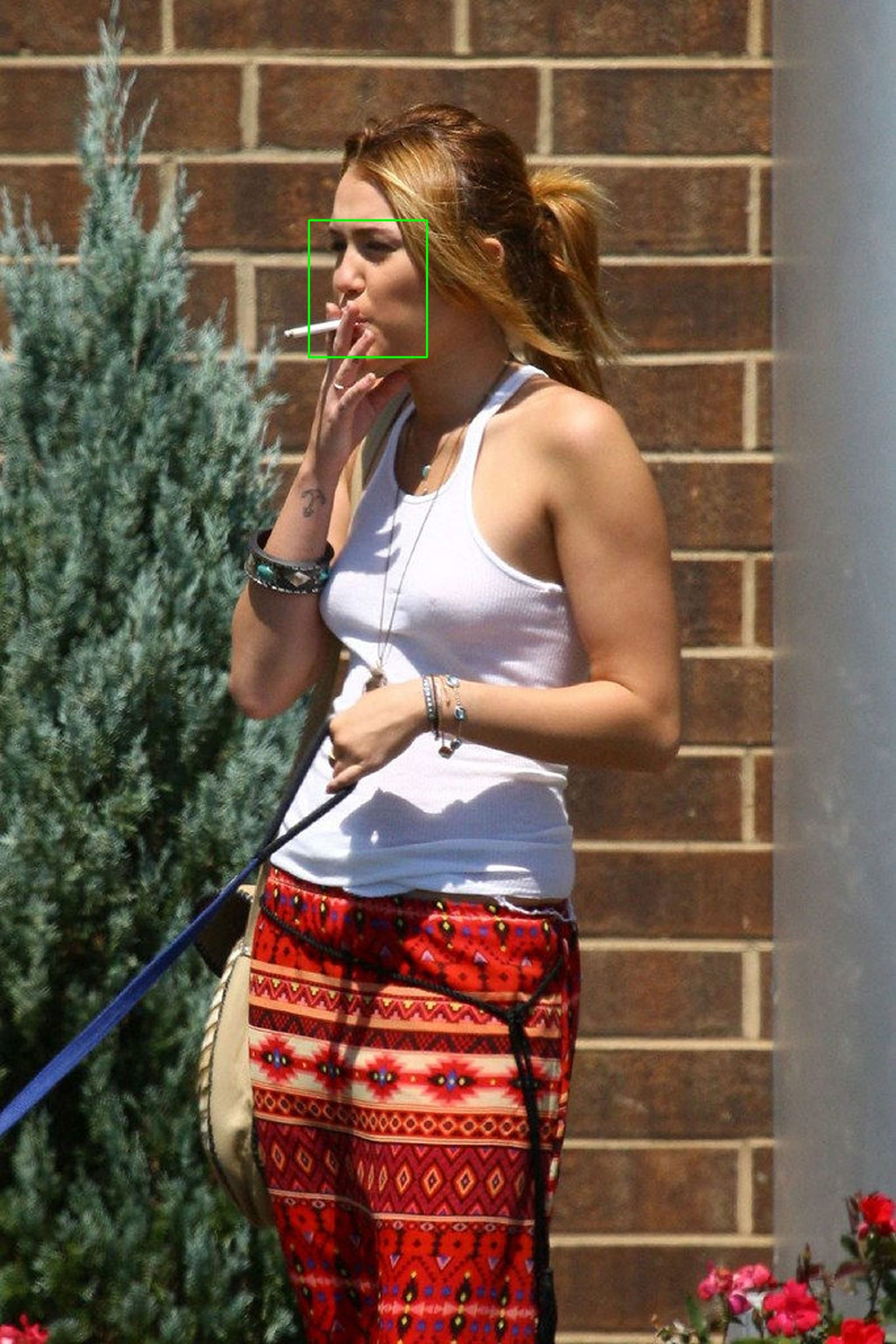} }}%
		\qquad
		\subfloat[\centering Adjusted bounding box ]{{\includegraphics[width=2cm,height=2.5cm]{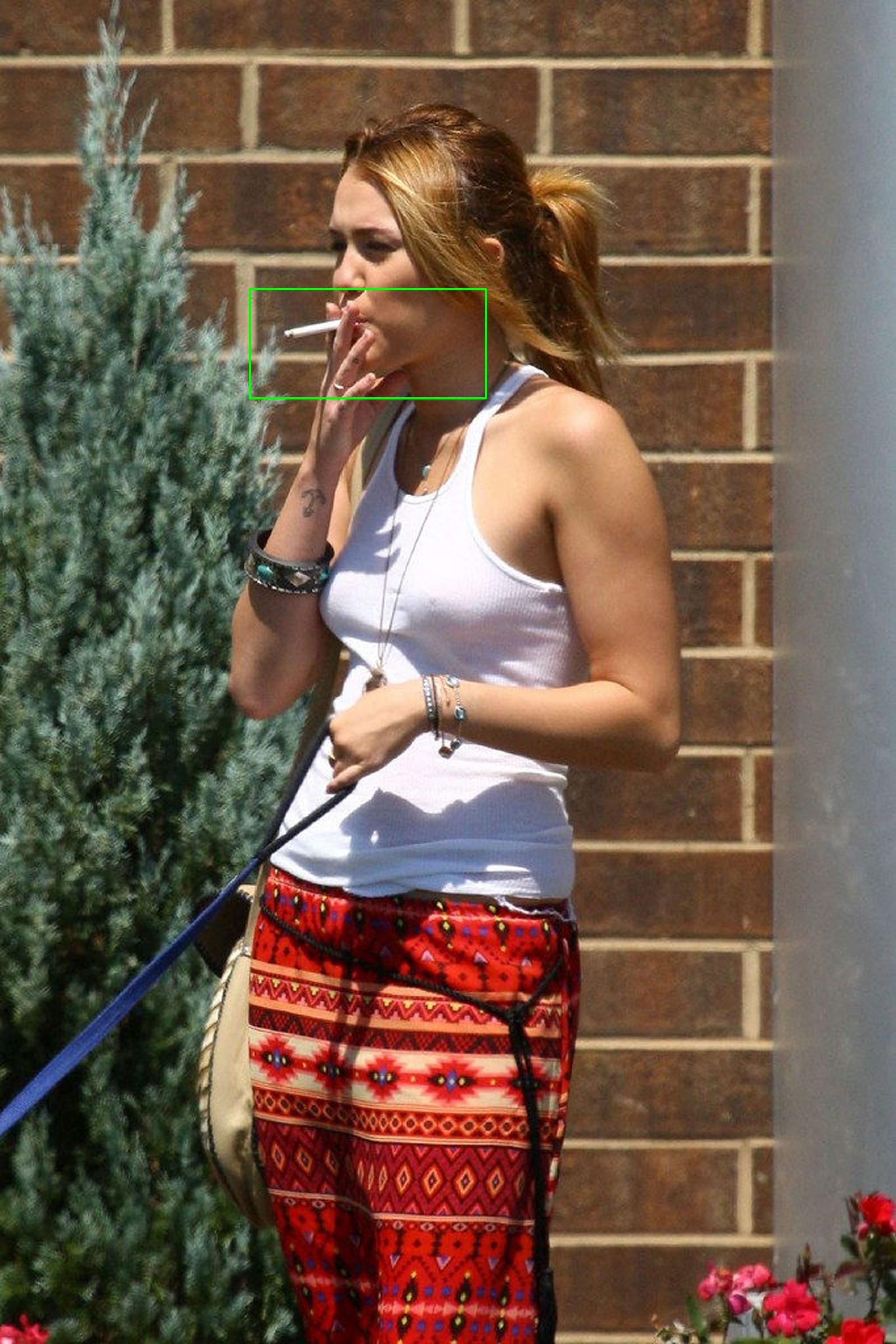} }}%
		\qquad
		\subfloat[\centering Extracted region of interest]{{\includegraphics[width=2cm,height=1.5cm]{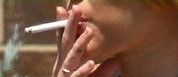} }}%
		\caption{Adjusted bounding box in Faced algorithm.\vspace{-.1in}}%
		\label{fig:adj}%
	\end{figure}

\begin{figure}[]%
		\centering
		\subfloat[\centering Original image]{{\includegraphics[width=2.2cm,height=2cm]{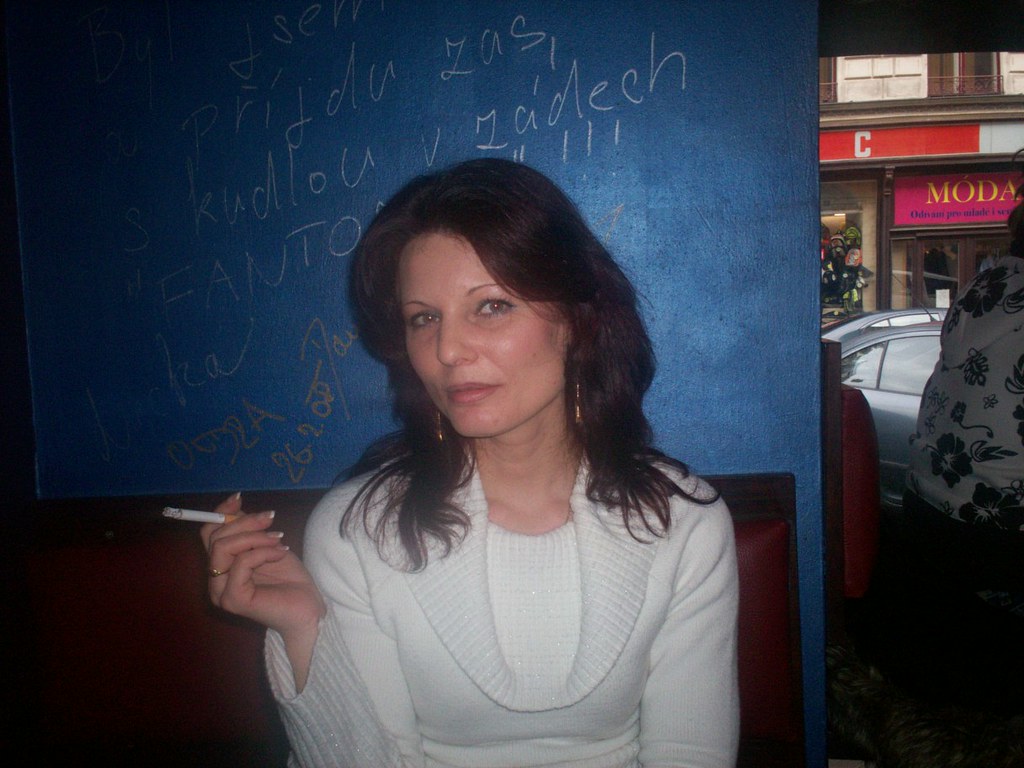} }}%
		\qquad
		\subfloat[\centering Hand detected by trained Yolo]{{\includegraphics[width=2.2cm,height=2cm]{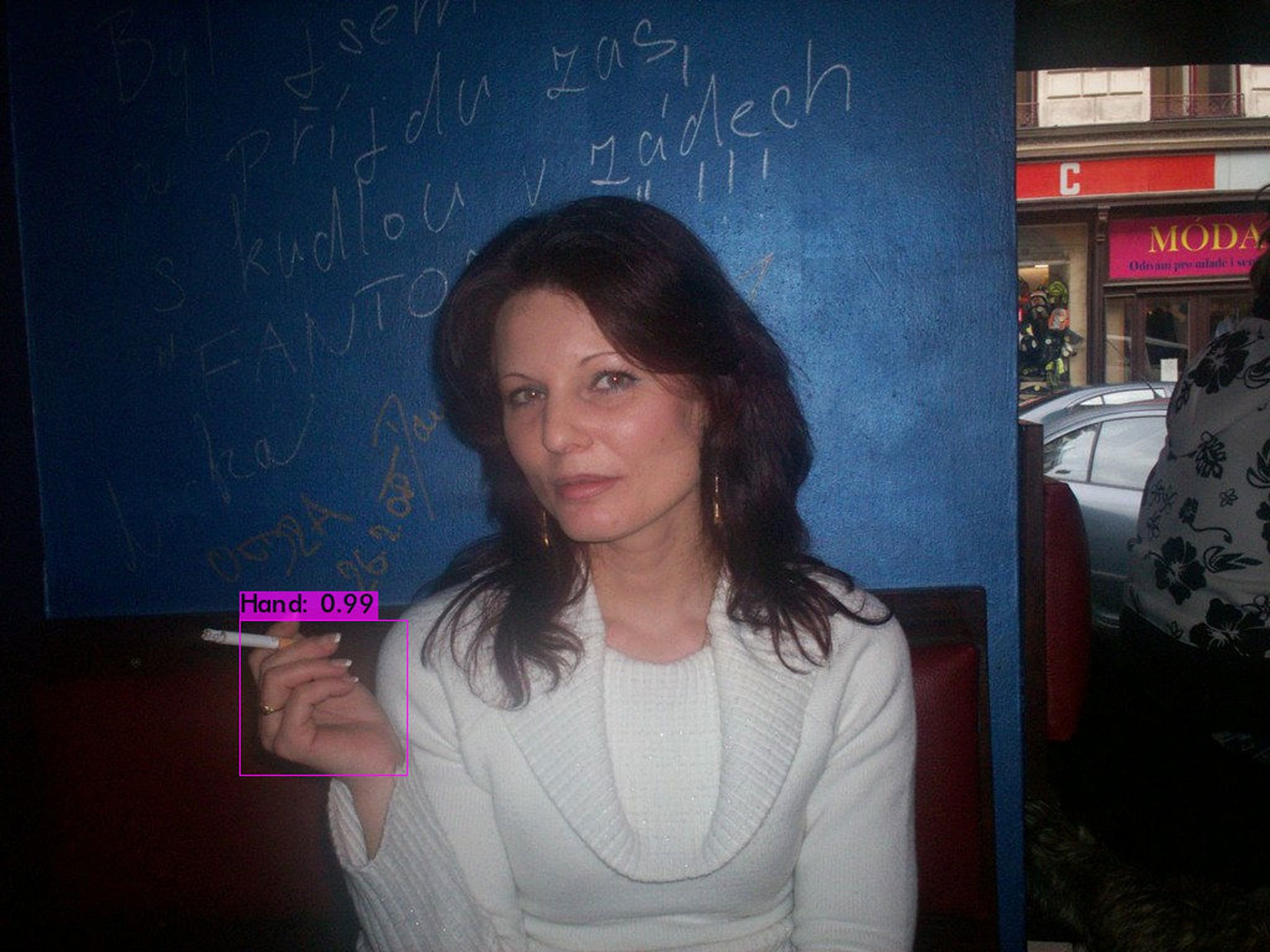} }}%
		\qquad
		\subfloat[\centering Extracted region of interest]{{\includegraphics[width=2cm,height=2cm]{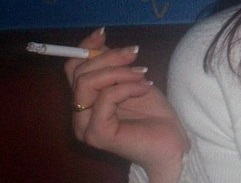} }}%
		\caption{Hand detection and region cropped by trained Yolo.\vspace{-.1in}}%
		\label{fig:crop}%
	\end{figure}

\vspace{-.08in}
	Due to the cigarette's small size, smoker and non-smoker objects look very similar in the broader view. This module performs data preprocessing, which solves challenges due to the small size of a cigarette. As shown in Fig.~\ref{fig:adj}~and~\ref{fig:crop}, the input images are preprocessed using \textit{Faced} algorithm (for detecting face regions)~\cite{r11} and Yolo-v3 (trained by us for detecting hand regions)~\cite{r10}, to extract the probable cigarette regions (i.e., face, hand). It helped us improve the model's performance since it needs to process relevant regions rather than process the whole image. We found \textit{Faced} algorithm more convincing on this dataset than the `Haar Cascade Classifier'~\cite{soo2014object} for extracting variable face poses. Further, we have fine-tuned the parameters of \textit{Faced} algorithm to improve the predictions.
	
For a given image, $I$, we have extracted face proposals $ F_1,F_2,.....,F_i, $ and hand proposals $ H_1,H_2,.....,H_j,$ where $i$ and $j$ are the number of proposals extracted by \textit{Faced} algorithm and trained yolo hand detector. \textit{Faced} algorithm returns the bounding box for a $k^{th}$ face proposal as per Eq.~\ref{eq:eq1} where, $c_x, c_y$ corresponds to coordinates of center and $w, h$ denotes width and height of the bounding box. \vspace{-.1in}
	\begin{equation} \label{eq:eq1}
	BBox_{F_k} = (c_x,c_y,w,h) 
	\end{equation}\vspace{-.28in}
	
Likewise, the Yolo hand detector returns the bounding box for $k^{th}$ hand proposal as per Eq.~\ref{eq:eq2} where $x_1, y_1$ denote the coordinates of top left corner and $x_2, y_2$ denote the coordinate of bottom right corner.\vspace{-.1in}
	\begin{equation} \label{eq:eq2}
	BBox_{H_k} = (x_1,y_1,x_2,y_2)
	\end{equation}\vspace{-.28in}
	
To improve the region’s coverage for a cigarette, we have adjusted the \textit{Faced} bounding box by shifting them vertically down with wider width so that cigarette orientation around the lips can be covered effectively in different cases. 
	Similarly, we have adjusted our trained Yolo-v3 for hand detection to ensure that the proposed regions can adequately cover the cigarette region.
	Adjustment in $k^{th}$ \textit{Faced} bounding box is performed as per Eq.~\ref{eq:eq3} where, $\delta_h$ and $\delta_v$ denote horizontal shift and vertical shift respectively.\vspace{-.1in}
	\begin{equation} \label{eq:eq3}
	Adj\_BBox_{F_k} = (c_x,c_y+\delta_v,w+\delta_h,h)      
	\end{equation}\vspace{-.25in}
	
	Adjustment in $k^{th}$ Yolo hand detector bounding box is performed as per Eq.~\ref{eq:eq4} where $\delta_h\ and\ \delta_v$ denote horizontal and vertical shift..\vspace{-.1in}
	\begin{equation} \label{eq:eq4}
	Adj\_BBox_{H_k} = (x_1-\delta_h,y_1-\delta_v,x_2+\delta_h,y_2+\delta_v) 
	\end{equation}\vspace{-.25in}
	
	\subsection{Classification Module}
	From baseline experiments in Section~\ref{sec:ablate}, it was observed that instead of using simple CNN and training from scratch, transfer learning can be used, which gives two-fold benefits in this problem by giving low-cost models and solving small dataset size issues. Here, we have used the ensemble of Resnet-18 and Resnet-34 models with their pre-trained weights. We have modified the Resnet model's architecture by classifying two classes in the fully-connected layer. We have used a softmax classifier and cross-entropy loss function.
	For any given image $I$, model performs classification on face and hand proposals as $C_{F_1},C_{F_2},...C_{F_i}$ and $C_{H_1},C_{H_2},...C_{H_j}$ repectively, where any $C_F, C_H$ is either $0$ or $1$. Finally classified category $C_I$ for the image $I$ is determined as per Eq.~\ref{eq:rq5} where $\oplus$ denotes the operation to takes maximum of all predicted classification categories.\vspace{-.1in}
	\begin{equation} \label{eq:rq5}
	C_I =  \oplus(C_{F_1},C_{F_2},...C_{F_i}, C_{H_1},C_{H_2},...C_{H_j}) 
	\end{equation}
	
	\subsection{Detection Module}
	This module implements real-time cigarette detection using Yolo-v3 trained on the cigarette images using LabelImg~\cite{r12}. This module is conditionally active and performs detection only if the given image is classified with smoking behavior. This idea helps in reducing false positives and improves performance significantly. The detection module gets triggered only when the given image, $I$ is classified as a smoker. Suppose $ind_i, ind_2, ......, ind_l$ denotes the indices of $l$ proposals on which $ C_F $, $ C_H $ have classified as a smoker. Then cigarette detector performs $l$ cigarette detections on the input image $I$. It is robust for various challenging situations (as shown in Fig.~\ref{fig:det1}). In case of failed detection in the first attempt, it performs the detection on $l$ proposals to improve our results (as shown in Fig.~\ref{fig:det2}). It overlays the proposals on the raw image to hold smokers' identities, which helps in the cases where cigarette is in hand. This idea is useful for cigarette monitoring systems.
	
	\section{EXPERIMENTS AND RESULTS} \label{sec:pagestyle}
	This section discusses the implementation details, evaluation metrics, and the results obtained during the experiments. \vspace{-.03in}
	
	\subsection{Implementation}
	\subsubsection{Experimental Setup}\vspace{-.02in}
	We have trained our proposed model on Nvidia RTX 2060 GPU having 1920 CUDA cores. This proposed model has been tested on Intel(R) Core(TM) i5-9300H, 2.40GHz, 16 GB RAM CPU machine with 64-bit Windows-10 OS machine.\vspace{-.04in}
	
	\subsubsection{Dataset and Training}\vspace{-.02in}
	The proposed model has been trained and evaluated on the Mendeley smoker dataset~\cite{dataset}, which has never been used before to the best of our knowledge. It contains 2,400 images with the smoker and non-smoker images in various poses and environmental settings. We have evaluated the proposed approach with the 80\%-20\% train-test split of dataset.\vspace{-.04in} 
	
	\subsubsection{Ablation Study}\label{sec:ablate}\vspace{-.02in}
	An ablation study to decide the proposed approach's architecture is performed and summarized in Table~\ref{tab:ablate}. Here, baselines are designed by considering the challenges due to the small dataset and cigarette size. We have used simple convolutional layers based architecture in baseline1 and baseline2. The importance of the region of interest (ROI) processing is shown by accuracy obtained in baseline1, which uses raw images compared to baseline2, which uses ROI. Baseline3, which uses raw images, is designed to evaluate the benefits of using transfer learning on this problem. Its architecture contains an ensemble of Resnet-18 and Resnet-34 models with their pre-trained weights~\cite{ensemble} and, when fed with raw images, gives better accuracy than baseline1 and 2. Based on the ablation study's results, it has been concluded that ROI processing and transfer learning help this problem. With the observations mentioned above, we came up with the proposed approach's architecture, as shown in Fig.~\ref{fig:visina8}. 
	
	\begin{table}[H]
		\centering
		\caption{Summar of the Ablation Study.\vspace{-.08in}}
		\label{tab:ablate}
		\begin{tabular}{lll}
			\hline
			\bf{Model} & \bf{Processing Strategy} & \bf{Accuracy} \\\hline
			Baseline1          & Raw input image              & 59\%     \\
			Baseline2          & Extracting ROIs     & 67\%     \\
			Baseline3    & Raw input image              & 90\%     \\
			Proposed Approach    & Extracting ROIs       & \bf{96.74}\%  \\\hline  
		\end{tabular}
	\end{table}

	\subsection{Results \& Evaluation}
	The proposed approach has obtained an accuracy of 96.74\%. Its performance has been evaluated using the following quantitative and qualitative measures.\vspace{-.2in}	
	\subsubsection{Quantitative Performance Measures}
	For our model's quantitative measure, we have shown its accuracy, precision, recall, and confusion matrix in Table~\ref{tab:quant}. \vspace{-.1in}
	
	\begin{table}[H]
		\centering
		\caption{Quantitative performance measures.\vspace{-.08in}}
		\label{tab:quant}
		\begin{tabular}{ll}
			\hline
			\bf{Metric}  & \bf{Obtained} \\\hline
			Precision       & 95\%              \\
			Recall          & 98\%              \\
			Accuracy        & 96.74\%              \\
			True Positives  & 197               \\
			True Negatives  & 190               \\
			False Positives & 10                \\
			False Negatives & 3                \\\hline  
		\end{tabular}\vspace{-.1in}
	\end{table}
	
	\subsubsection{Qualitative Performance Measures}
\begin{figure}[]
		\centering
		\includegraphics[width=8.8cm, height=2.5cm]{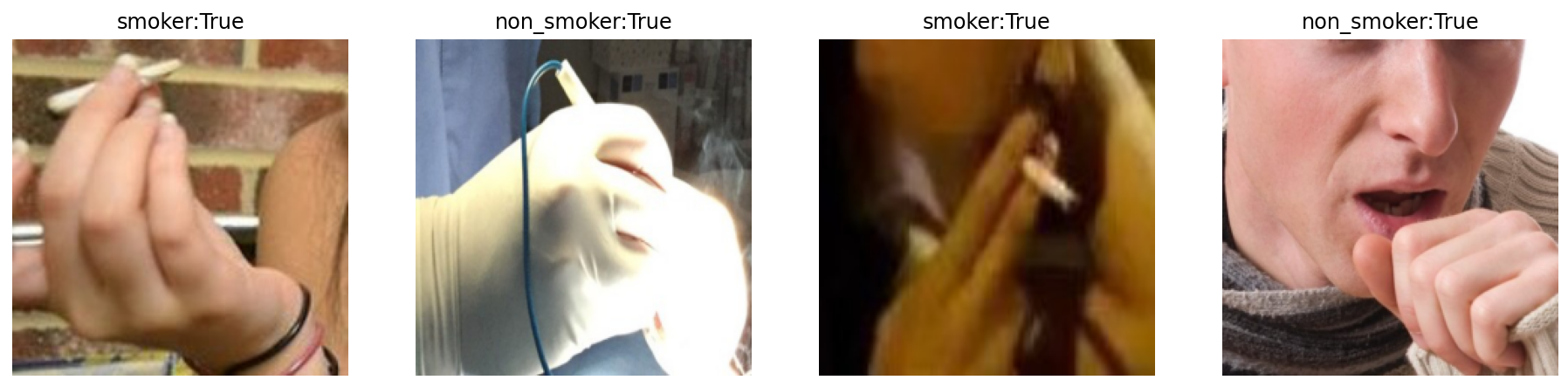}
		\caption{Classification on sample images.\vspace{-.1in}}\label{fig:res_visina8}
	\end{figure}

\begin{figure}[H]
		\centering
		\subfloat[\centering Dark background ]{{\includegraphics[width=3.5cm,height=2.5cm,keepaspectratio]{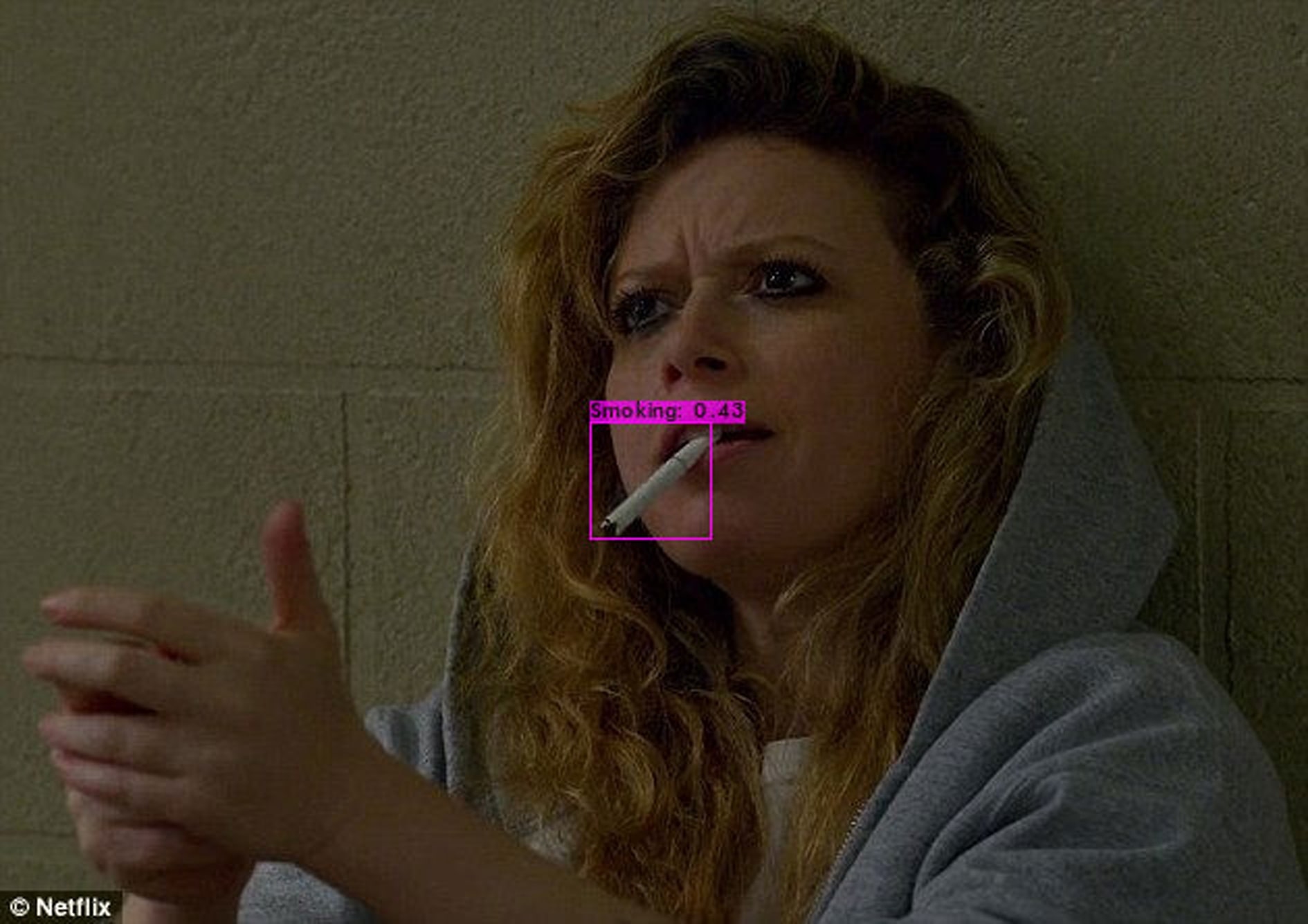} }}%
		\qquad
		\subfloat[\centering Side face pose ]{{\includegraphics[width=3.5cm,height=2.5cm,keepaspectratio]{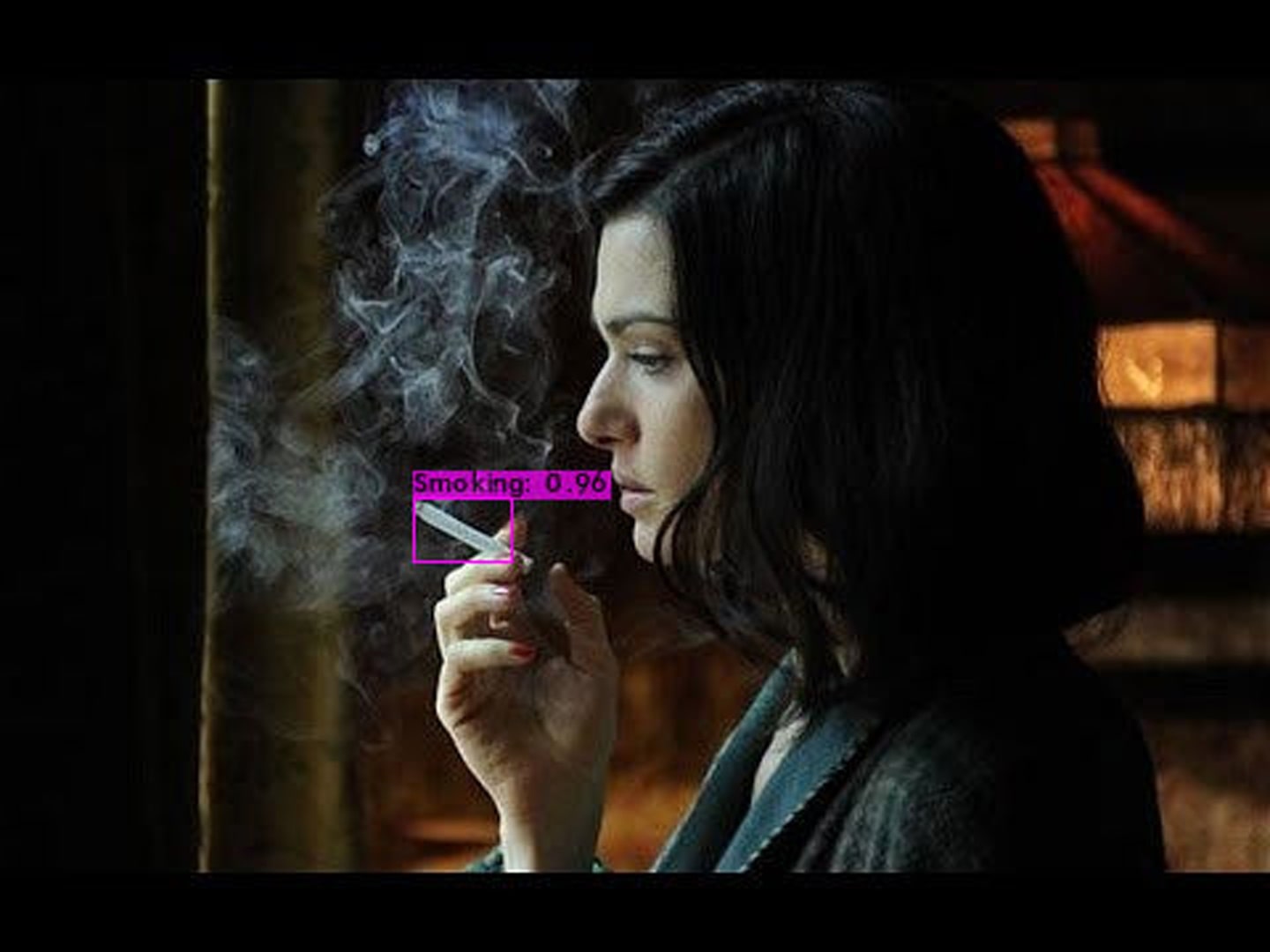} }}%
		\qquad
		\subfloat[\centering Small visibility of cigerette]{{\includegraphics[width=3.5cm,height=2.5cm, keepaspectratio]{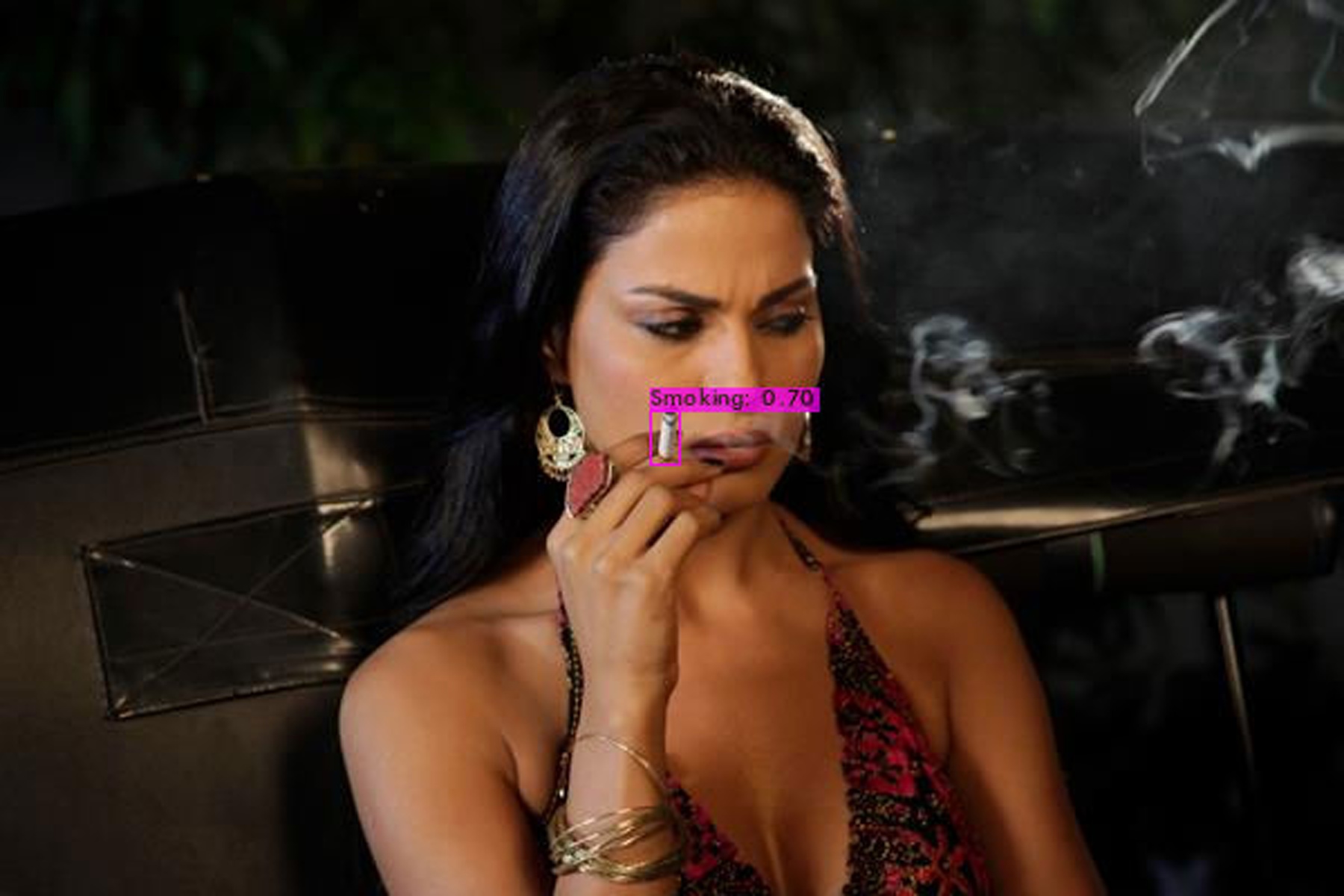} }}%
		\qquad
		\subfloat[\centering Multi-subject enviornment ]{{\includegraphics[width=3.5cm,height=2.5cm,keepaspectratio]{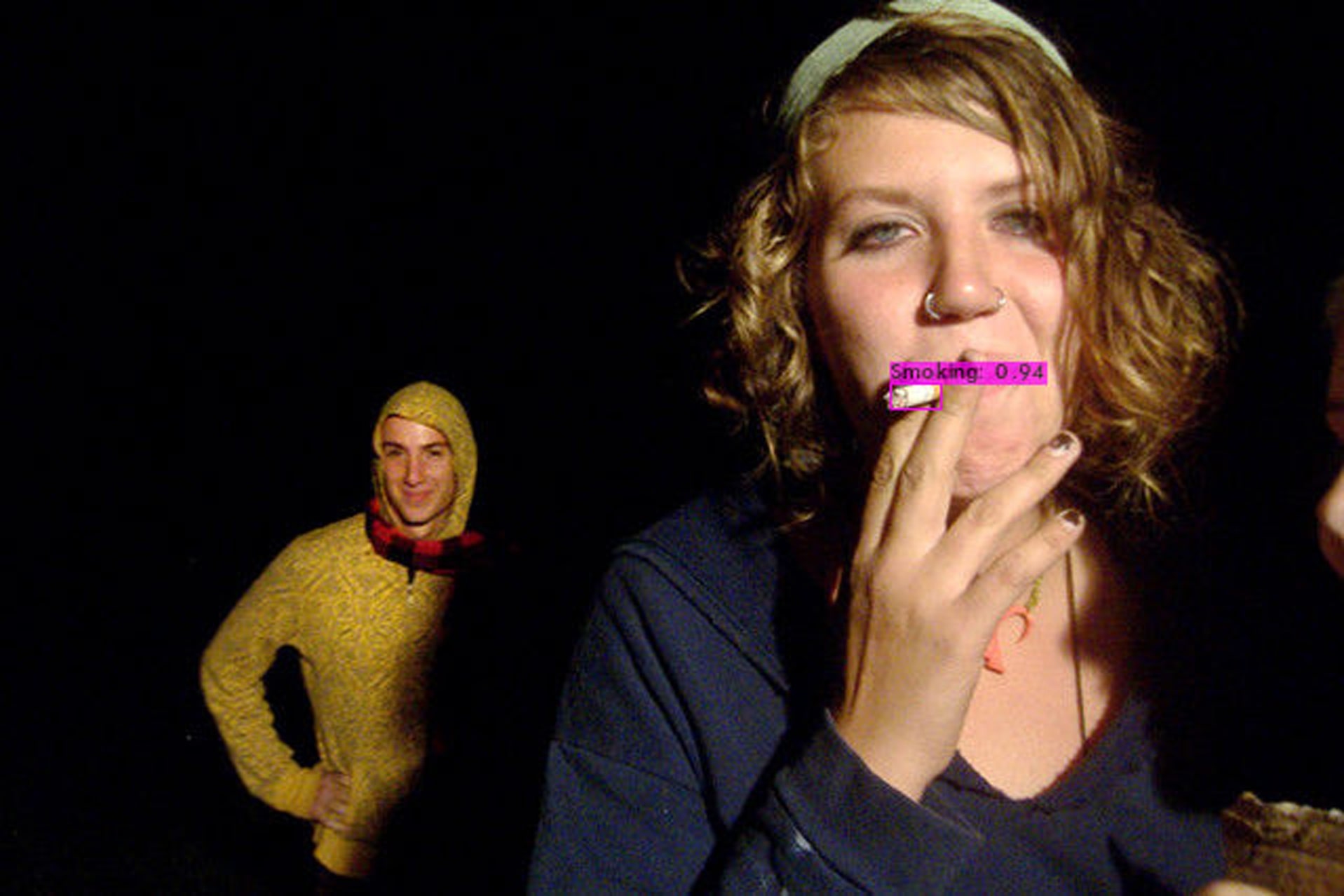} }}%
		\qquad
		\caption{Cigarette detection in different settings.\vspace{-.1in}}
		\label{fig:det1}%
	\end{figure}

\begin{figure}[H]
		\centering
		\subfloat[\centering No cigarette detected]{{\includegraphics[width=2.7cm,height=2.5cm]{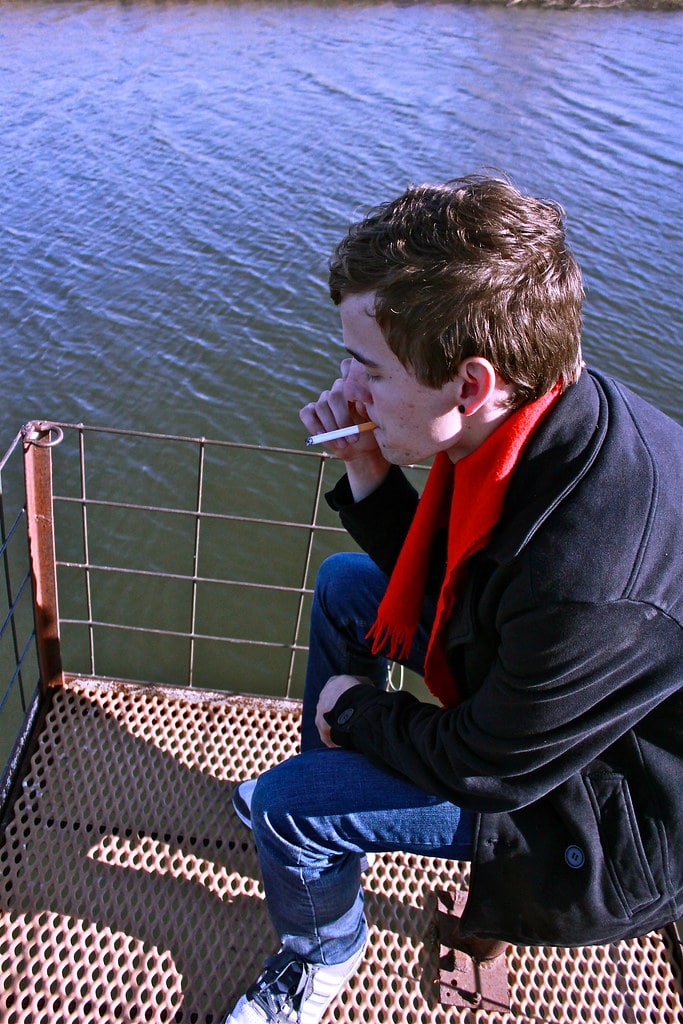} }}%
		\qquad
		\subfloat[\centering Cigarette detected]{{\includegraphics[width=3cm,height=2.5cm]{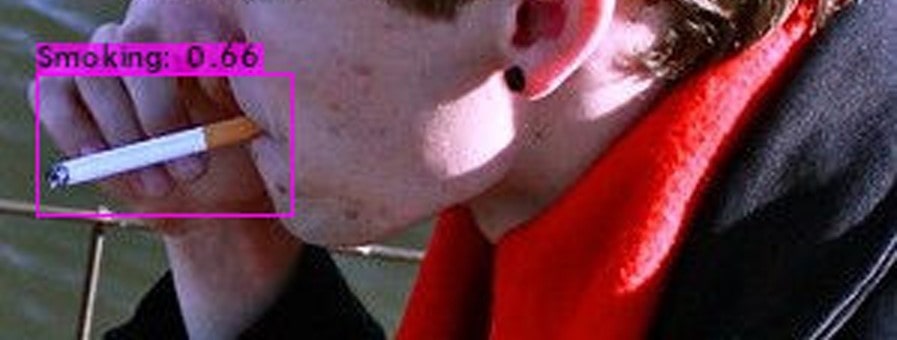} }}%
		\qquad
		\subfloat[\centering False detection]{{\includegraphics[width=2.7cm,height=2.5cm]{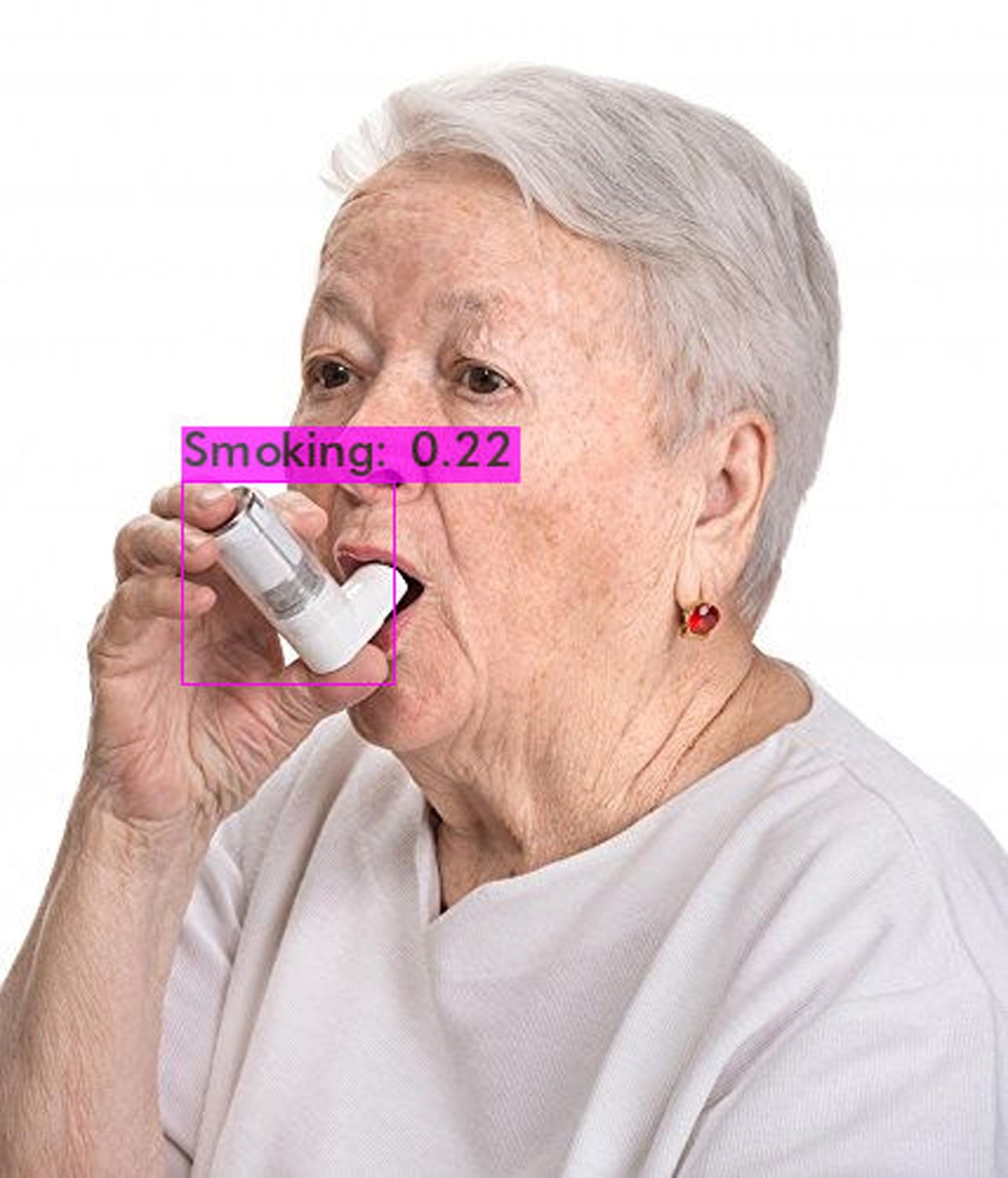} }}%
		\qquad
		\subfloat[\centering False detection removed ]{{\includegraphics[width=2.9cm,height=2.5cm]{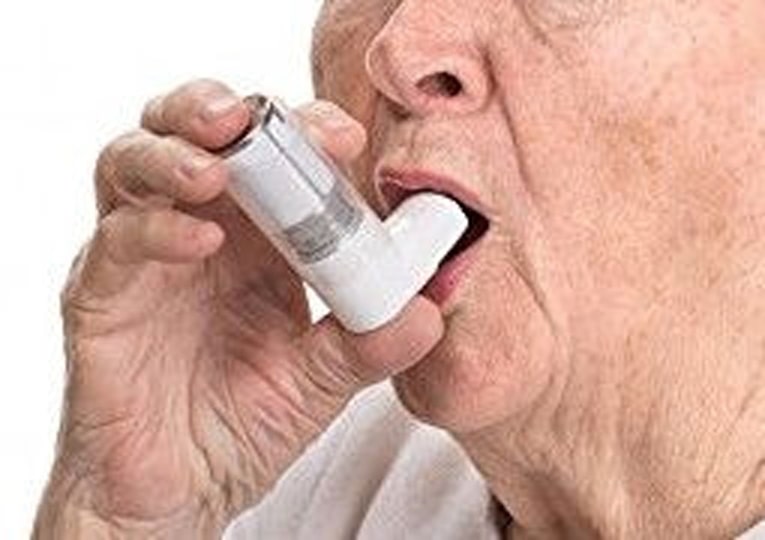} }}%
		\qquad	
		\caption{Improvement in detection using proposed approach.\vspace{-.1in}}%
		\label{fig:det2}%
	\end{figure}

	The classification results are shown in Fig.~\ref{fig:res_visina8} while the detection results are shown in Fig.~\ref{fig:det1} and Fig.~\ref{fig:det2} in different challenging scenarios.\vspace{-.1in}
	
	\subsection{Comparison with the state-of-the-art approaches}
	As per the literature, no state-of-the-art (SOTA) approaches for cigarette usage analysis are available for the Mendeley smoker dataset~\cite{dataset}. Moreover, few datasets are available for this problem. We have compared the proposed approach to SOTA approaches for other problems with similar objectives and use-cases. The comparison shown in Table~\ref{tab:sota} affirms the proposed approach's applicability for this problem. \vspace{-.05in}
	
	\begin{table}[H]
		\centering
		\caption{Comparison with state-of-the-art approaches\vspace{-.05in}}
		\label{tab:sota}
		\begin{tabular}{lll}
			\hline
			\bf{Method} & \bf{Author} & \bf{Accuracy} \\\hline
			CNN Based   & {Ou et al.}         \cite{sota3}     & {79.4\%}\\
			Deep Learning          & {Dhanwal et al.} \cite{e08}              & {89.9\%}     \\
			Wrist IMU          & {A\~nazco et al.} \cite{e05}              & {91.38\%}     \\
			Faster-RCNN          & {Lu et al.} \cite{e07}              & {92.1\%}     \\\Xhline{.025\arrayrulewidth}
			
			\multicolumn{2}{c}{Proposed} & \bf{96.74}\%\\\hline
		\end{tabular}
	\end{table}
	
	\section{CONCLUSION AND FUTURE WORK} \label{sec:typestyle}
	This paper has proposed a novel approach for smoking behavior classification and detection using deep learning. The proposed approach has obtained significant results in challenging situations for the Mendeley smoker dataset, which has never been used so far. It can be extended for anomalous human activity recognition and detection of other small objects. In the future, we aim to make the proposed approach more effective by including the information from more modalities such as videos and feeds through night-vision cameras.\vspace{-.05in}
	
	\section*{Acknowledgements}\vspace{-.1in}
	This work was supported by the University Grants Commission (UGC) INDIA with grant number: 190510040512.
	
	\bibliographystyle{IEEEbib}
	\bibliography{CigDetect}
	
\end{document}